\documentclass[10pt]{article}

\usepackage[margin=1in]{geometry}
\usepackage{amsmath,amssymb,amsfonts}
\usepackage{booktabs}
\usepackage{multirow}
\usepackage{graphicx}
\usepackage{microtype}
\usepackage{xcolor}
\usepackage{hyperref}
\usepackage[round,authoryear]{natbib}
\usepackage{algorithm}
\usepackage{algorithmic}
\usepackage{enumitem}

\hypersetup{
    colorlinks=true,
    linkcolor=blue,
    citecolor=blue,
    urlcolor=blue
}

\title{Graph-Operator World Models for Morphology-Parameter Generalization in Continuous Control}

\author{Xu Yang$^1$, Yiqin Yang$^2$, Qianchuan Zhao$^1$ \\
$^1$Tsinghua University \\
$^2$The Key Laboratory of Cognition and Decision Intelligence for Complex Systems, \\
Institute of Automation, Chinese Academy of Sciences 
}
\date{}

\newcommand{\G}{\mathcal{G}}
\newcommand{\D}{\mathcal{D}}

\newcommand{\R}{\mathbb{R}}
\newcommand{\stopgrad}{\operatorname{sg}}

\begin{document}
\maketitle

\begin{abstract}
World models for continuous control are commonly trained for a fixed physical system and can degrade when known morphology parameters such as link lengths, masses, damping, and actuation change.
Existing approaches often provide these parameters as conditioning information, but leave unspecified which part of the learned transition should remain reusable and which part should change with morphology.
We propose \emph{Graph-Operator World Models} (GraphOp-WM), a structured world model for generalization across unseen morphology parameters within related articulated robot families.
GraphOp-WM represents bodies and their kinematic relations as an attributed graph and factorizes each transition into a morphology-independent local dynamics basis and a morphology-conditioned structured operator.
The operator combines node-local modulation, kinematic-tree coupling, and a low-rank global correction, while architectural information separation, basis normalization, and paired-morphology supervision encourage static morphology dependence to be carried by the operator pathway.
Graph-level readout and edge-wise action representations provide a compatible interface for reward, value, and TD-MPC-style planning.
We further define controlled MuJoCo parameter splits covering interpolation, extrapolation, and held-out compositions of link geometry, mass, damping, and actuation parameters in Hopper, Walker2d, and HalfCheetah.
\end{abstract}

\section{Introduction}

World models support sample-efficient reinforcement learning by predicting latent transitions, rewards, and values for planning or imagined policy optimization.
DreamerV3 \citep{hafner2025dreamerv3}, TD-MPC2 \citep{hansen2024tdmpc2}, and PWM \citep{georgiev2025pwm} demonstrate that learned dynamics models can support strong control across broad task suites.
Their generality is primarily obtained through model capacity, shared task embeddings, and multitask data.
Recent morphology-conditioned world models additionally encode robot specifications or structural embeddings, showing that known embodiment information can improve dynamics modeling under embodiment variation \citep{danesh2026qwm,wang2026westworld,he2025cross}.
However, conditioning morphology as an input does not by itself specify which components of the transition should be invariant across parameter changes and which components should account for morphology-induced changes in dynamical coupling.

Robotic dynamics additionally expose explicit compositional structure.
Even within a fixed articulated topology, changing link geometry, body mass, inertia, joint damping, or actuator gear changes how local motion propagates through the robot.
These changes alter the transition field while preserving substantial reusable local structure.
A generic context vector can inform a dynamics model that the morphology has changed, but it does not prescribe how morphology parameters should transform the coupling between local dynamic responses.

We study the following question:
\begin{quote}
Can a compact world model generalize to unseen physical morphology parameters by explicitly separating reusable local dynamics from morphology-induced coupling?
\end{quote}

We introduce GraphOp-WM, a graph-structured world model designed around this separation.
Bodies are represented as nodes, while joints and actuators are represented as edges.
A shared local kernel predicts a canonical increment for each body using only local dynamic state, local action information, and a discrete body type.
It is deliberately prevented from observing masses, lengths, inertias, morphology identifiers, or pooled morphology embeddings.
A separate graph encoder converts the MuJoCo morphology graph into a structured transition operator that acts on the shared local increments.

For body-token matrix $Z_t\in\R^{N_m\times d}$, the transition is
\begin{equation}
    \hat Z_{t+1}=Z_t+\mathcal{A}_{m,t}\bar B_t,
    \label{eq:transition_intro}
\end{equation}
where $N_m$ is fixed within a robot family, $\bar B_t$ is a normalized shared local dynamics basis, and $\mathcal{A}_{m,t}$ is a morphology-parameter-conditioned graph operator.
The operator is structured as
\begin{equation}
    \mathcal{A}_{m,t}
    =I+\operatorname{Diag}(d_{m,t})
    +E_m\operatorname{Diag}(w_{m,t})E_m^\top
    +U_{m,t}\operatorname{Diag}(s_{m,t})U_{m,t}^\top .
    \label{eq:operator_intro}
\end{equation}
The diagonal component captures body-local response, the incidence-matrix component follows the kinematic tree, and the low-rank term captures residual global modes such as trunk--limb and contact-mediated coupling.
We do not claim that this operator is an exact inverse mass matrix.
It is a constrained transition parameterization motivated by the locality and repeated structure of articulated dynamics.

GraphOp-WM uses graph-level readout for reward and value prediction and attaches actions to actuator edges.
This preserves the physical organization of the robot in the predictive and control interfaces without requiring morphology parameters to be collapsed into a single unconstrained context vector.

Our contributions are:
\begin{itemize}
    \item We formulate morphology-parameter generalization for world models: predicting and planning under unseen changes in physical parameters while keeping the articulated topology fixed within each robot family.
    \item We introduce a transition factorization that separates a morphology-independent local dynamics basis from a structured operator conditioned on static morphology parameters, combining local, graph-supported, and low-rank global coupling.
    \item We impose architectural information separation, basis normalization, and paired-morphology supervision so that morphology-dependent transition changes are assigned to the operator rather than absorbed by the shared basis.
    \item We specify controlled MuJoCo splits for interpolation, extrapolation, and held-out compositions of geometry, mass, damping, and actuation parameters, enabling direct measurement of parameter-efficient out-of-distribution dynamics and planning generalization.
\end{itemize}

\section{Related Work}

\paragraph{World models for continuous control.}
PETS \citep{chua2018pets} and MBPO \citep{janner2019mbpo} learn probabilistic dynamics for planning or policy optimization.
DreamerV3 learns recurrent latent dynamics and trains behavior through imagined trajectories \citep{hafner2025dreamerv3}.
TD-MPC combines a task-oriented latent dynamics model with short-horizon planning and terminal value estimation \citep{hansen2022tdmpc}.
TD-MPC2 uses a decoder-free latent model and short-horizon planning across a large continuous-control suite \citep{hansen2024tdmpc2}.
PWM pretrains multitask world models and extracts policies through first-order gradients \citep{georgiev2025pwm}.
GraphOp-WM retains the control-centric modeling and planning interface of TD-MPC, but introduces an explicit factorization of reusable local dynamics and morphology-parameter-dependent coupling.

\paragraph{Contextual and morphology-conditioned world models.}
CaDM infers a dynamics context from recent transitions before conditioning a forward model on that context \citep{lee2020cadm}, while MAMBA uses recurrent world-model states to represent task information over a meta-episode \citep{rimon2024mamba}.
These approaches are suited to latent properties that must be inferred from interaction.
Robot geometry, inertial parameters, and joint connectivity are instead available before interaction in MJCF, URDF, or USD specifications.
QWM conditions a Dreamer-style model on normalized quadruped specifications \citep{danesh2026qwm}, and WestWorld combines structural embeddings with system-aware mixture-of-experts routing across heterogeneous robotic systems \citep{wang2026westworld}.
A particle-based graph world model provides a shared state and action interface across dexterous embodiments \citep{he2025cross}.
GraphOp-WM targets a narrower but more explicit question: for known changes in physical morphology parameters, it treats morphology as a structured operator over local dynamic responses rather than only as a pooled vector, token embedding, or expert-routing signal. This yields a concrete inductive bias for extrapolating to unseen parameter values and combinations.

\paragraph{Morphology-conditioned control.}
NerveNet represents an articulated agent as a graph and shares policy computation across body parts \citep{wang2018nervenet}.
Shared Modular Policies instantiate one reusable module per actuator and coordinate modules through messages along the morphology \citep{huang2020smp}.
MetaMorph treats morphology as a conditioning modality for a Transformer controller \citep{gupta2022metamorph}.
Body Transformer restricts attention using a sensor--actuator graph \citep{sferrazza2025bot}, and URMA uses morphology-agnostic encoders and decoders for multi-embodiment locomotion \citep{bohlinger2025urma}.
These works focus primarily on transferring control computation across bodies or embodiments, whereas GraphOp-WM applies graph structure to factorize predictive dynamics under morphology-parameter variation.

\paragraph{Graph dynamics and learned operators.}
Interaction Networks and graph-network simulators establish object- and relation-centric inductive biases for physical prediction and control \citep{battaglia2016interaction,sanchezgonzalez2018gn}.
Contrastively learned structured world models use graph transitions over object-centric latent states \citep{kipf2020cswm}.
Compositional Koopman operators combine graph encoders with block-structured linear dynamics for systems with variable object counts \citep{li2020koopman}.
MS-HGNN further incorporates kinematic structure and morphological symmetry into robotic dynamics learning \citep{xie2025mshgnn}.
GraphOp-WM factorizes prediction into a morphology-independent local basis and a graph-generated operator.
The graph-supported term follows the kinematic tree explicitly, while a low-rank component represents residual nonlocal coupling.

\section{Problem Setting}

We consider a family of control problems indexed by morphology $m$:
\begin{equation}
    \mathcal{M}_m=(\mathcal S_m,\mathcal A_m,T_m,r_k,\Omega_m,\rho_{0,m},\gamma),
\end{equation}
where state and action dimensions may vary with the robot, while $k$ indexes the control objective.
Each morphology is represented by an attributed graph
\begin{equation}
    \G_m=(\mathcal V_m,\mathcal E_m^{\mathrm{joint}},\mathcal E_m^{\mathrm{act}}).
\end{equation}
Body nodes $i\in\mathcal V_m$ contain geometry and inertial metadata.
Joint edges contain kinematic relations, and actuator edges associate commands with controlled joints.

The training set contains trajectories from morphologies $m\in\mathcal M_{\mathrm{train}}$.
The primary objective is zero-shot dynamics prediction and planning on unseen variants in $\mathcal M_{\mathrm{test}}$ from related articulated families.
The benchmark separates interpolation, parameter extrapolation, held-out parameter combinations, and within-family topology extension.
Universal transfer to arbitrary robot families remains outside the scope of the formulation.

\section{Method}

\subsection{Morphology Graph}

We parse MuJoCo XML or DMControl specifications into a body--joint graph.
For body $i$, static node attributes are
\begin{equation}
    p_i^v=[m_i,I_i,\mathrm{geom}_i,\mathrm{type}_i,\mathrm{contact}_i,\mathrm{frame}_i].
\end{equation}
For joint edge $(i,j)$, static attributes are
\begin{equation}
    p_{ij}^e=[\mathrm{jointtype}_{ij},\mathrm{axis}_{ij},\mathrm{origin}_{ij},
    \mathrm{range}_{ij},\mathrm{damping}_{ij},\mathrm{gear}_{ij}].
\end{equation}
Dynamic body features include local position, orientation, linear and angular velocity, and contact signals.
Joint position, joint velocity, and actuator command are attached to the corresponding edge.

Graphs are batched by concatenating nodes and edges and storing a graph-membership index, as in standard packed graph processing.
Within each robot family, the graph topology and the correspondence between bodies, joints, and actuators are fixed.
Every body token has fixed channel dimension $d$; morphology variation is represented through the static node and edge attributes rather than changes in graph size.

\subsection{State Tokens and Global Readout}

A shared body encoder produces dynamic node tokens
\begin{equation}
    z_{t,i}=E_\phi(x_{t-L:t,i}^v,x_{t-L:t,ij}^e,\mathrm{type}_i)\in\R^d.
\end{equation}
The encoder may use a short recurrent or attention history for partial observations.
It does not receive masses, lengths, inertias, morphology identifiers, or pooled morphology embeddings.
This blocks direct access to static morphology through the state pathway.

For scalar graph-level predictions, a learned global query attends to all body tokens:
\begin{align}
    \alpha_{t,i}
    &=\operatorname{softmax}_i
    \left(\frac{q_g^\top W_kz_{t,i}}{\sqrt d}\right),\\
    h_t^g
    &=\sum_{i=1}^{N_m}\alpha_{t,i}W_vz_{t,i}.
\end{align}
We concatenate this readout with the root-body token:
\begin{equation}
    g_t=[z_{t,\mathrm{root}};h_t^g]\in\R^{2d}.
    \label{eq:global_readout}
\end{equation}
Thus reward and value heads have fixed input dimension without fixing the number of bodies.

\subsection{Morphology-Independent Local Dynamics Basis}

For each body, the shared local kernel predicts an unconstrained candidate increment using local state, incident dynamic joint features, and local actuator commands:
\begin{equation}
    B_{t,i}
    =b_\theta
    \left(
    z_{t,i},
    \operatorname{Agg}_{j\in\mathcal N(i)}[q_{t,ij},\dot q_{t,ij},a_{t,ij}],
    \mathrm{type}_i
    \right).
    \label{eq:local_basis}
\end{equation}
The kernel is shared across all bodies and morphologies and has no access to static morphology parameters.
To remove the scale ambiguity between the basis and operator, we normalize each graph's basis by its root-mean-square magnitude:
\begin{equation}
    \bar B_t
    =\frac{B_t}{\sqrt{\frac{1}{N_md}\|B_t\|_F^2+\epsilon}}.
    \label{eq:basis_norm}
\end{equation}

\subsection{Structured Morphology Operator}

A morphology encoder processes only static graph attributes and returns node coefficients, edge coefficients, and $r$ node-aligned global modes:
\begin{equation}
    (d_m,w_m,\tilde U_m,s_m)=G_\psi(\G_m).
\end{equation}
Here $d_m\in\R^{N_m}$, $w_m\in\R^{|\mathcal E_m^{\mathrm{joint}}|}$, $\tilde U_m\in\R^{N_m\times r}$, and $s_m\in\R^r$.
The same node and edge encoders are reused for every graph, so their parameter dimensions do not depend on $N_m$.
We orthogonalize the global modes within each graph:
\begin{equation}
    U_m=\operatorname{qr}(\tilde U_m),\qquad U_m^\top U_m=I_{r_m},
\end{equation}
where $r_m=\min(r,N_m)$ is the effective rank for morphology $m$.

The static coefficients are modulated by shared state-dependent gates:
\begin{align}
    d_{m,t,i}&=d_{m,i}\tanh(g_d(z_{t,i})),\\
    w_{m,t,ij}&=w_{m,ij}\tanh(g_e(z_{t,i},z_{t,j},x_{t,ij}^e)),\\
    s_{m,t}&=s_m\odot\tanh(g_s(h_t^g)).
\end{align}
The gates contain no trainable morphology identifiers; they only activate the static structure according to the current pose, velocity, and contact state.

Let $E_m\in\R^{N_m\times |\mathcal E_m^{\mathrm{joint}}|}$ be the oriented incidence matrix of the kinematic graph.
The transition operator is
\begin{equation}
    \mathcal A_{m,t}
    =I
    +\operatorname{Diag}(d_{m,t})
    +E_m\operatorname{Diag}(w_{m,t})E_m^\top
    +U_m\operatorname{Diag}(s_{m,t})U_m^\top .
    \label{eq:structured_operator}
\end{equation}
The predicted transition is
\begin{equation}
    \hat Z_{t+1}=Z_t+\mathcal A_{m,t}\bar B_t.
    \label{eq:graph_transition}
\end{equation}

Equation~\eqref{eq:structured_operator} is not presented as an exact rigid-body solver.
Its components encode three inductive biases: local response, coupling supported by the kinematic tree, and a small number of nonlocal residual modes.
The low-rank term is retained only if numerical transition-Jacobian diagnostics show a rapidly decaying residual spectrum after removing local and graph-supported components.

\subsection{Information-Separated Factorization}

The architecture enforces information separation: static morphology attributes enter only $\mathcal A_{m,t}$, while the local basis observes only dynamic local state and action.
Basis normalization removes arbitrary reciprocal scaling between $\bar B_t$ and $\mathcal A_{m,t}$.
We additionally use paired-morphology transitions within each robot family.

For two variants initialized at an aligned normalized pose and driven by the same normalized action sequence, a single local basis is used with two operators:
\begin{align}
    \widehat{\Delta Z}_t^{m_1}&=\mathcal A_{m_1,t}\bar B_t,\\
    \widehat{\Delta Z}_t^{m_2}&=\mathcal A_{m_2,t}\bar B_t.
\end{align}
The paired loss is
\begin{equation}
    \mathcal L_{\mathrm{pair}}
    =\left\|\widehat{\Delta Z}_t^{m_1}-\Delta Z_t^{m_1}\right\|_2^2
    +\left\|\widehat{\Delta Z}_t^{m_2}-\Delta Z_t^{m_2}\right\|_2^2.
    \label{eq:pair_loss}
\end{equation}
This supervision assigns morphology-dependent transition differences to the operator while retaining a shared local basis.
Operator-swap analysis uses the same dynamic state and basis with operators from controlled morphology variants and compares the induced transition changes with simulator rollouts.

\subsection{Reward, Value, and Variable-Action Heads}

Robot dynamics are shared across control objectives, but reward semantics may be task dependent.
We therefore allow a fixed-dimensional goal or task embedding $e_k$ to enter reward, value, and policy heads, but never the transition basis or morphology operator.

The reward is decomposed into global and local components:
\begin{equation}
    \hat r_t
    =r_\eta^g(g_t,e_k)
    +\sum_{i=1}^{N_m}r_\eta^v(z_{t,i})
    +\sum_{e\in\mathcal E_m^{\mathrm{act}}}r_\eta^a(z_{t,e},a_{t,e}).
    \label{eq:reward}
\end{equation}
Here $z_{t,e}$ is an actuator-edge representation computed from its incident body tokens and static joint attributes.
The global term captures progress, height, and uprightness; the local sums capture body contact and actuator effort.
For diagnostic experiments, known MuJoCo reward functions can be evaluated directly from predicted physical quantities, isolating transition quality from reward-model error.

The value head uses the fixed-dimensional readout:
\begin{equation}
    V_\xi(\mathcal Z_t,k)=V_\xi(g_t,e_k).
\end{equation}
For action values, actions are embedded on actuator edges, followed by an action-conditioned graph readout:
\begin{equation}
    Q_\omega(\mathcal Z_t,A_t,k)
    =Q_\omega(\operatorname{Readout}_Q(Z_t,\G_m,A_t),e_k).
\end{equation}
Both heads share weight dimensions across arbitrary body and actuator counts.

When a policy prior is used, each actuated edge independently predicts a Gaussian action conditioned on its incident body tokens and the global readout:
\begin{equation}
    (\mu_{t,e},\sigma_{t,e})
    =\pi_\rho(z_{t,i},z_{t,j},p_{ij}^e,g_t,e_k),
    \qquad e=(i,j)\in\mathcal E_m^{\mathrm{act}}.
    \label{eq:edge_policy}
\end{equation}
The number of output actions therefore equals the number of actuators in the current morphology.

\subsection{Training Objective}

The primary dynamics loss predicts token-aligned physical quantities or anchored latent targets:
\begin{equation}
    \mathcal L_{\mathrm{dyn}}
    =\sum_{h=1}^{H_{\mathrm{train}}}\rho^{h-1}
    \frac{1}{N_m}
    \left\|\hat Z_{t+h}-Z_{t+h}^{\mathrm{tar}}\right\|_F^2,
\end{equation}
where the loss is evaluated only over nodes present in each packed graph.
For proprioceptive experiments, an observation head anchors position, velocity, root motion, and contact channels.

The value target is a standard bootstrapped TD target rather than self-consistency between two unconstrained value predictions:
\begin{align}
    y_t&=r_t+\gamma(1-d_t)\stopgrad V_{\bar\xi}(\mathcal Z_{t+1},k),\\
    \mathcal L_V&=\left\|V_\xi(\mathcal Z_t,k)-y_t\right\|_2^2.
\end{align}

The operator is regularized around the identity and by the magnitude of its structured corrections:
\begin{equation}
    \mathcal L_{\mathrm{op}}
    =\lambda_d\|d_{m,t}\|_2^2
    +\lambda_w\|w_{m,t}\|_2^2
    +\lambda_s\|s_{m,t}\|_1.
\end{equation}
The complete objective is
\begin{equation}
    \mathcal L
    =\mathcal L_{\mathrm{dyn}}
    +\lambda_r\mathcal L_r
    +\lambda_Q\mathcal L_Q
    +\lambda_V\mathcal L_V
    +\lambda_{\mathrm{pair}}\mathcal L_{\mathrm{pair}}
    +\lambda_{\mathrm{op}}\mathcal L_{\mathrm{op}}.
    \label{eq:total_loss}
\end{equation}

\begin{algorithm}[t]
\caption{Training GraphOp-WM}
\begin{algorithmic}[1]
\STATE Sample morphology-parameter variants and trajectory segments from $\D$.
\STATE Pack body nodes, joint edges, actuator edges, and graph-membership indices.
\STATE Encode dynamic body tokens $Z_t$ without static morphology attributes.
\STATE Predict and normalize the shared local basis $\bar B_t$.
\STATE Encode static morphology coefficients from $\G_m$ and apply state-dependent gates.
\STATE Construct $\mathcal A_{m,t}$ using Eq.~\eqref{eq:structured_operator}.
\STATE Predict $\hat Z_{t+1}$ using Eq.~\eqref{eq:graph_transition} and roll forward for $H_{\mathrm{train}}$ steps.
\STATE Compute dynamics, reward, TD, paired-morphology, and operator losses.
\STATE Update model parameters and target value parameters.
\end{algorithmic}
\end{algorithm}

\section{Model-Based Control with Variable Graphs}

TD-MPC-style planning does not conceptually require a fixed latent or action dimension.
At a decision step, all candidate trajectories correspond to the current morphology, whose node and actuator counts are known.
For morphology $m$, MPC samples
\begin{equation}
    A_{t:t+K-1}
    \in\R^{K\times N_{\mathrm{sample}}\times |\mathcal E_m^{\mathrm{act}}|},
\end{equation}
replicates the current graph state across candidates, and rolls out GraphOp-WM.
The planning objective is
\begin{equation}
    A_{t:t+K-1}^\star
    =\arg\max_A
    \sum_{k=0}^{K-1}\gamma^k\hat r_{t+k}
    +\gamma^K V_{\bar\xi}(\hat{\mathcal Z}_{t+K},k).
\end{equation}
Different morphologies instantiate different action tensors at separate planning calls; no global maximum action dimension is required.

To adapt TD-MPC2, we replace its vector encoder and MLP transition with packed graph modules, replace reward and value inputs with Eq.~\eqref{eq:global_readout}, and replace the fixed policy output with Eq.~\eqref{eq:edge_policy}.
The MPPI/CEM optimization logic and temporal-difference targets remain unchanged.

\section{MuJoCo Benchmark Definition}

\subsection{Environment Families and Morphology Construction}

The benchmark uses the MuJoCo physics engine \citep{todorov2012mujoco} and three locomotion families: Hopper, Walker2d, and HalfCheetah.
Each canonical MJCF model is converted into the attributed graph in Section~4.1 and then transformed by deterministic MJCF edits.
Parameter variants scale link geometry, body mass, torso mass, joint damping, and actuator gear relative to the canonical model.
Body inertias are recomputed from the scaled geometry and mass rather than varied independently.
Ground friction, gravity, integration time step, termination rules, and reward definitions remain fixed within each family.

\subsection{Parameter Splits}

All values in Table~\ref{tab:morphology_splits} are multipliers of the canonical MJCF quantity.
Training morphologies are generated by maximin Latin-hypercube sampling over the training ranges.
Interpolation morphologies use unseen values and combinations inside the training envelope.
Extrapolation morphologies lie outside that envelope.
The compositional split combines individually observed directions into jointly unseen morphologies.

\begin{table*}[t]
\centering
\caption{MuJoCo morphology parameterization. Values are multiplicative factors relative to each canonical Hopper, Walker2d, or HalfCheetah model. Inertia is deterministically recomputed after geometry and mass changes.}
\label{tab:morphology_splits}
\small
\setlength{\tabcolsep}{5pt}
\resizebox{\textwidth}{!}{%
\begin{tabular}{p{2.5cm}p{3.0cm}p{3.0cm}p{3.2cm}p{4.0cm}}
\toprule
\textbf{Morphology quantity} &
\textbf{Training envelope} &
\textbf{Interpolation set} &
\textbf{Extrapolation set} &
\textbf{Held-out composition} \\
\midrule
Link length &
$[0.80,\,1.20]$ &
$\{0.85,0.95,1.05,1.15\}$ &
$[0.60,0.75]\cup[1.25,1.40]$ &
distal link $1.25$ with actuator gear $0.70$ \\
Link mass &
$[0.75,\,1.25]$ &
$\{0.85,0.95,1.05,1.15\}$ &
$[0.50,0.70]\cup[1.30,1.50]$ &
distal mass $1.35$ with torso mass $0.75$ \\
Torso mass &
$[0.80,\,1.20]$ &
$\{0.90,1.00,1.10\}$ &
$[0.60,0.75]\cup[1.25,1.40]$ &
torso mass $1.30$ with limb mass $0.70$ \\
Joint damping &
$[0.70,\,1.30]$ &
$\{0.80,0.90,1.10,1.20\}$ &
$[0.40,0.60]\cup[1.40,1.60]$ &
damping $1.40$ with actuator gear $0.70$ \\
Actuator gear &
$[0.80,\,1.20]$ &
$\{0.90,0.95,1.05,1.10\}$ &
$[0.60,0.75]\cup[1.25,1.40]$ &
gear $0.70$ with distal link length $1.25$ \\
Body inertia &
recomputed &
recomputed &
recomputed &
recomputed \\
\bottomrule
\end{tabular}
}
\end{table*}

\section{Conclusion}

We introduced GraphOp-WM, a structured world model for morphology-parameter generalization in continuous control.
The method represents an articulated robot as a graph and separates a morphology-independent local dynamics basis from a structured operator conditioned on physical morphology parameters.
Architectural information separation, basis normalization, and paired-morphology supervision assign reusable and morphology-dependent components distinct roles in the transition.
The accompanying MuJoCo benchmark defines interpolation, extrapolation, and held-out composition splits for studying out-of-distribution prediction and planning under changes in geometry, mass, damping, and actuation.
GraphOp-WM is intended as a compact structural prior for parameter generalization within related articulated robot families rather than a universal cross-topology controller.

\bibliographystyle{plainnat}
\bibliography{references}

@inproceedings{chua2018pets,
  title     = {Deep Reinforcement Learning in a Handful of Trials Using Probabilistic Dynamics Models},
  author    = {Chua, Kurtland and Calandra, Roberto and McAllister, Rowan and Levine, Sergey},
  booktitle = {Advances in Neural Information Processing Systems},
  volume    = {31},
  year      = {2018},
  url       = {https://proceedings.neurips.cc/paper/2018/hash/3de568f8597b94bda53149c7d7f5958c-Abstract.html}
}

@inproceedings{janner2019mbpo,
  title     = {When to Trust Your Model: Model-Based Policy Optimization},
  author    = {Janner, Michael and Fu, Justin and Zhang, Marvin and Levine, Sergey},
  booktitle = {Advances in Neural Information Processing Systems},
  volume    = {32},
  year      = {2019},
  url       = {https://proceedings.neurips.cc/paper/2019/hash/5faf461eff3099671ad63c6f3f094f7f-Abstract.html}
}

@article{hafner2025dreamerv3,
  title   = {Mastering Diverse Control Tasks through World Models},
  author  = {Hafner, Danijar and Pasukonis, Jurgis and Ba, Jimmy and Lillicrap, Timothy},
  journal = {Nature},
  volume  = {640},
  pages   = {647--653},
  year    = {2025},
  doi     = {10.1038/s41586-025-08744-2},
  url     = {https://doi.org/10.1038/s41586-025-08744-2}
}

@inproceedings{hansen2022tdmpc,
  title     = {Temporal Difference Learning for Model Predictive Control},
  author    = {Hansen, Nicklas A. and Su, Hao and Wang, Xiaolong},
  booktitle = {Proceedings of the 39th International Conference on Machine Learning},
  series    = {Proceedings of Machine Learning Research},
  volume    = {162},
  pages     = {8387--8406},
  year      = {2022},
  publisher = {PMLR},
  url       = {https://proceedings.mlr.press/v162/hansen22a.html}
}

@inproceedings{hansen2024tdmpc2,
  title     = {{TD-MPC2}: Scalable, Robust World Models for Continuous Control},
  author    = {Hansen, Nicklas and Su, Hao and Wang, Xiaolong},
  booktitle = {International Conference on Learning Representations},
  year      = {2024},
  url       = {https://openreview.net/forum?id=FzpfPa6unv},
  eprint    = {2310.16828},
  archivePrefix = {arXiv}
}

@inproceedings{georgiev2025pwm,
  title     = {{PWM}: Policy Learning with Multi-Task World Models},
  author    = {Georgiev, Ignat and Giridhar, Varun and Hansen, Nicklas and Garg, Animesh},
  booktitle = {International Conference on Learning Representations},
  year      = {2025},
  url       = {https://openreview.net/forum?id=hOELrZfg0J},
  eprint    = {2407.02466},
  archivePrefix = {arXiv}
}

@inproceedings{lee2020cadm,
  title     = {Context-aware Dynamics Model for Generalization in Model-Based Reinforcement Learning},
  author    = {Lee, Kimin and Seo, Younggyo and Lee, Seunghyun and Lee, Honglak and Shin, Jinwoo},
  booktitle = {Proceedings of the 37th International Conference on Machine Learning},
  series    = {Proceedings of Machine Learning Research},
  volume    = {119},
  pages     = {5757--5766},
  year      = {2020},
  publisher = {PMLR},
  url       = {https://proceedings.mlr.press/v119/lee20g.html}
}

@inproceedings{rimon2024mamba,
  title     = {{MAMBA}: An Effective World Model Approach for Meta-Reinforcement Learning},
  author    = {Rimon, Zohar and Jurgenson, Tom and Krupnik, Orr and Adler, Gilad and Tamar, Aviv},
  booktitle = {International Conference on Learning Representations},
  year      = {2024},
  url       = {https://openreview.net/forum?id=1RE0H6mU7M},
  eprint    = {2403.09859},
  archivePrefix = {arXiv}
}

@inproceedings{wang2018nervenet,
  title     = {{NerveNet}: Learning Structured Policy with Graph Neural Networks},
  author    = {Wang, Tingwu and Liao, Renjie and Ba, Jimmy and Fidler, Sanja},
  booktitle = {International Conference on Learning Representations},
  year      = {2018},
  url       = {https://openreview.net/forum?id=S1sqHMZCb}
}

@inproceedings{huang2020smp,
  title     = {One Policy to Control Them All: Shared Modular Policies for Agent-Agnostic Control},
  author    = {Huang, Wenlong and Mordatch, Igor and Pathak, Deepak},
  booktitle = {Proceedings of the 37th International Conference on Machine Learning},
  series    = {Proceedings of Machine Learning Research},
  volume    = {119},
  pages     = {4455--4464},
  year      = {2020},
  publisher = {PMLR},
  url       = {https://proceedings.mlr.press/v119/huang20d.html}
}

@inproceedings{gupta2022metamorph,
  title     = {{MetaMorph}: Learning Universal Controllers with Transformers},
  author    = {Gupta, Agrim and Fan, Linxi and Ganguli, Surya and Fei-Fei, Li},
  booktitle = {International Conference on Learning Representations},
  year      = {2022},
  url       = {https://openreview.net/forum?id=Opmqtk_GvYL},
  eprint    = {2203.11931},
  archivePrefix = {arXiv}
}

@inproceedings{sferrazza2025bot,
  title     = {Body Transformer: Leveraging Robot Embodiment for Policy Learning},
  author    = {Sferrazza, Carmelo and Huang, Dun-Ming and Liu, Fangchen and Lee, Jongmin and Abbeel, Pieter},
  booktitle = {Proceedings of The 8th Conference on Robot Learning},
  series    = {Proceedings of Machine Learning Research},
  volume    = {270},
  pages     = {3407--3424},
  year      = {2025},
  publisher = {PMLR},
  url       = {https://proceedings.mlr.press/v270/sferrazza25a.html}
}

@inproceedings{bohlinger2025urma,
  title     = {One Policy to Run Them All: An End-to-End Learning Approach to Multi-Embodiment Locomotion},
  author    = {Bohlinger, Nico and Czechmanowski, Grzegorz and Krupka, Maciej Piotr and Kicki, Piotr and Walas, Krzysztof and Peters, Jan and Tateo, Davide},
  booktitle = {Proceedings of The 8th Conference on Robot Learning},
  series    = {Proceedings of Machine Learning Research},
  volume    = {270},
  pages     = {3356--3378},
  year      = {2025},
  publisher = {PMLR},
  url       = {https://proceedings.mlr.press/v270/bohlinger25a.html}
}

@inproceedings{battaglia2016interaction,
  title     = {Interaction Networks for Learning about Objects, Relations and Physics},
  author    = {Battaglia, Peter W. and Pascanu, Razvan and Lai, Matthew and Rezende, Danilo Jimenez and Kavukcuoglu, Koray},
  booktitle = {Advances in Neural Information Processing Systems},
  volume    = {29},
  year      = {2016},
  url       = {https://proceedings.neurips.cc/paper/2016/hash/3147da8ab4a0437c15ef51a5cc7f2dc4-Abstract.html}
}

@inproceedings{sanchezgonzalez2018gn,
  title     = {Graph Networks as Learnable Physics Engines for Inference and Control},
  author    = {Sanchez-Gonzalez, Alvaro and Heess, Nicolas and Springenberg, Jost Tobias and Merel, Josh and Riedmiller, Martin and Hadsell, Raia and Battaglia, Peter},
  booktitle = {Proceedings of the 35th International Conference on Machine Learning},
  series    = {Proceedings of Machine Learning Research},
  volume    = {80},
  pages     = {4470--4479},
  year      = {2018},
  publisher = {PMLR},
  url       = {https://proceedings.mlr.press/v80/sanchez-gonzalez18a.html}
}

@inproceedings{kipf2020cswm,
  title     = {Contrastive Learning of Structured World Models},
  author    = {Kipf, Thomas and van der Pol, Elise and Welling, Max},
  booktitle = {International Conference on Learning Representations},
  year      = {2020},
  url       = {https://openreview.net/forum?id=H1gax6VtDB},
  eprint    = {1911.12247},
  archivePrefix = {arXiv}
}

@inproceedings{li2020koopman,
  title     = {Learning Compositional Koopman Operators for Model-Based Control},
  author    = {Li, Yunzhu and He, Hao and Wu, Jiajun and Katabi, Dina and Torralba, Antonio},
  booktitle = {International Conference on Learning Representations},
  year      = {2020},
  url       = {https://openreview.net/forum?id=H1ldzA4tPr},
  eprint    = {1910.08264},
  archivePrefix = {arXiv}
}

@inproceedings{xie2025mshgnn,
  title     = {Morphological-Symmetry-Equivariant Heterogeneous Graph Neural Network for Robotic Dynamics Learning},
  author    = {Xie, Fengze and Wei, Sizhe and Song, Yue and Yue, Yisong and Gan, Lu},
  booktitle = {Proceedings of the 7th Annual Learning for Dynamics \& Control Conference},
  series    = {Proceedings of Machine Learning Research},
  volume    = {283},
  pages     = {1392--1405},
  year      = {2025},
  publisher = {PMLR},
  url       = {https://proceedings.mlr.press/v283/xie25a.html}
}

@misc{danesh2026qwm,
  title         = {Toward Hardware-Agnostic Quadrupedal World Models via Morphology Conditioning},
  author        = {Danesh, Mohamad H. and Li, Chenhao and Abyaneh, Amin and Houssaini, Anas and Ellis, Kirsty and Berseth, Glen and Hutter, Marco and Lin, Hsiu-Chin},
  year          = {2026},
  eprint        = {2604.08780},
  archivePrefix = {arXiv},
  primaryClass  = {cs.RO},
  url           = {https://arxiv.org/abs/2604.08780}
}

@misc{wang2026westworld,
  title         = {{WestWorld}: A Knowledge-Encoded Scalable Trajectory World Model for Diverse Robotic Systems},
  author        = {Wang, Yuchen and Kong, Jiangtao and Wei, Sizhe and Li, Xiaochang and Lin, Haohong and Zhao, Hongjue and Zhou, Tianyi and Gan, Lu and Shao, Huajie},
  year          = {2026},
  eprint        = {2603.14392},
  archivePrefix = {arXiv},
  primaryClass  = {cs.RO},
  url           = {https://arxiv.org/abs/2603.14392}
}

@misc{he2025cross,
  title         = {Scaling Cross-Embodiment World Models for Dexterous Manipulation},
  author        = {He, Zihao and Ai, Bo and Mu, Tongzhou and Liu, Yulin and Wan, Weikang and Fu, Jiawei and Du, Yilun and Christensen, Henrik I. and Su, Hao},
  year          = {2025},
  eprint        = {2511.01177},
  archivePrefix = {arXiv},
  primaryClass  = {cs.RO},
  url           = {https://arxiv.org/abs/2511.01177}
}

@inproceedings{todorov2012mujoco,
  title     = {{MuJoCo}: A Physics Engine for Model-Based Control},
  author    = {Todorov, Emanuel and Erez, Tom and Tassa, Yuval},
  booktitle = {2012 IEEE/RSJ International Conference on Intelligent Robots and Systems},
  pages     = {5026--5033},
  year      = {2012},
  publisher = {IEEE},
  doi       = {10.1109/IROS.2012.6386109},
  url       = {https://doi.org/10.1109/IROS.2012.6386109}
}




\end{document}